\documentclass[9pt, conference]{IEEEtran}
\IEEEoverridecommandlockouts
\DeclareUnicodeCharacter{2217}{*}
\usepackage{cite}
\usepackage{amsmath,amssymb,amsfonts}
\usepackage{algorithmic}
\usepackage{booktabs}
\usepackage{graphicx}
\usepackage{textcomp}
\usepackage{xcolor}
\usepackage{multirow}
\usepackage{algorithm}
\usepackage{multirow}
\usepackage{amsmath}
\usepackage{hyperref}
\usepackage{balance}
\def\BibTeX{{\rm B\kern-.05em{\sc i\kern-.025em b}\kern-.08em
    T\kern-.1667em\lower.7ex\hbox{E}\kern-.125emX}}

\begin{document}

\title{DeepCircuitX: A Comprehensive Repository-Level Dataset for RTL Code Understanding, Generation, and PPA Analysis}

% \author{
% {Zeju Li}$^{1\dagger}$ \quad Changran Xu$^{1\dagger}$ \quad Zhengyuan Shi$^{1\dagger}$ \quad Zedong Peng$^2$ \quad Yi Liu$^1$ \quad Yunhao Zhou$^2$ \\ 
% {Lingfeng Zhou}$^3$ \quad {Chengyu Ma}$^4$ \quad {Jianyuan Zhong}$^1$ \quad {Xi Wang}$^5$ \quad {Jieru Zhao}$^2$ \quad {Zhufei Chu}$^4$ \\ 
% {Xiaoyan Yang}$^3$ \quad {Qiang Xu}$^{1*}$\thanks{$\dagger$ Equal Contribution, *Corresponding author, qxu@cse.cuhk.edu.hk.} \\[6pt]
% $^1$The Chinese University of Hong Kong \quad $^2$Shanghai Jiao Tong University \\[3pt]
% $^3$Hangzhou Dianzi University \quad $^4$Ningbo University \quad $^5$Southeast University
% }

\author{
	\IEEEauthorblockN{
        Zeju Li $^{1,6\dagger}$, 
        Changran Xu $^{1,6\dagger}$, 
        Zhengyuan Shi $^{1,6\dagger}$, 
        Zedong Peng $^{2,6}$, 
        Yi Liu $^{1,6}$, 
        Yunhao Zhou $^{1,6}$, 
        Lingfeng Zhou $^{3,6}$, \\
        Chengyu Ma $^{4,6}$, 
        Jianyuan Zhong $^{1,6}$, 
        Xi Wang $^{5,6}$, 
        Jieru Zhao $^{2}$, 
        Zhufei Chu $^{4}$, 
        Xiaoyan Yang $^{3}$, 
        Qiang Xu $^{1,6}$ \thanks{$\dagger$ Equal Contribution} \thanks{Corresponding author: Qiang Xu (qxu@cse.cuhk.edu.hk)}} 

\IEEEauthorblockA{$^1$\textit{Department of Computer Science and Engineering}, \textit{The Chinese University of Hong Kong}, Sha Tin, Hong Kong S.A.R.\\}
\IEEEauthorblockA{$^2$ \textit{Department of Computer Science and Engineering}, Shanghai, China \\}
\IEEEauthorblockA{$^3$\textit{School of Computer Science}, \textit{Hangzhou Dianzi University}, Hangzhou, China \\}
\IEEEauthorblockA{$^4$\textit{Faculty of Electrical Engineering and Computer Science}, \textit{Ningbo University}, Ningbo, China \\}
\IEEEauthorblockA{$^5$ \textit{School of Integrated Circuit, Southeast University}, Nanjing, China \\}
\IEEEauthorblockA{$^6$\textit{National Center of Technology Innovation for EDA}, Nanjing, China \\}
} 

\maketitle

\begin{abstract}
This paper introduces DeepCircuitX, a comprehensive repository-level dataset designed to advance RTL (Register Transfer Level) code understanding, generation, and power-performance-area (PPA) analysis. Unlike existing datasets that are limited to either file-level RTL code or physical layout data, DeepCircuitX provides a holistic, multilevel resource that spans repository, file, module, and block-level RTL code. This structure enables more nuanced training and evaluation of large language models (LLMs) for RTL-specific tasks. DeepCircuitX is enriched with Chain of Thought (CoT) annotations, offering detailed descriptions of functionality and structure at multiple levels. These annotations enhance its utility for a wide range of tasks, including RTL code understanding, generation, and completion. Additionally, the dataset includes synthesized netlists and PPA metrics, facilitating early-stage design exploration and enabling accurate PPA prediction directly from RTL code. We demonstrate the dataset's effectiveness on various LLMs finetuned with our dataset and confirm the quality with human evaluations. Our results highlight DeepCircuitX as a critical resource for advancing RTL-focused machine learning applications in hardware design automation. Our data is available at \url{https://zeju.gitbook.io/lcm-team}.

\end{abstract}

\begin{figure*}[!t]
    \centering
    \includegraphics[width=0.8\linewidth]{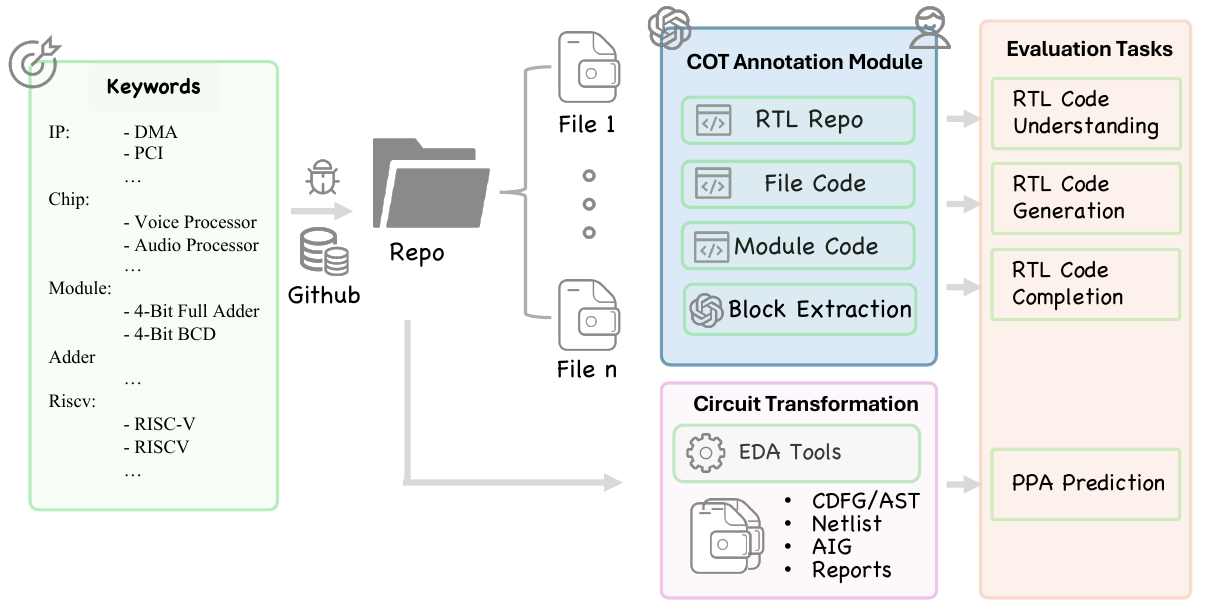}
    \caption{Pipeline overview of the proposed framework, illustrating the key stages: data collection from GitHub using keywords, data annotation via chain-of-thought (COT), circuit transformation, and evaluation, including RTL code tasks for LLM and PPA prediction.}
    \label{fig:overview}
\end{figure*}

\section{Introduction} \label{Sec:Intro}

Register Transfer Level (RTL) modeling is a crucial step in the Electronic Design Automation (EDA) flow, where RTL code (e.g., Verilog and VHDL) represents hardware functionality, bridging design specifications and circuit implementations. Engineers then translate RTL into netlists, floorplans, and layouts, directly influencing circuit design quality.
Integrating Artificial Intelligence (AI) into the EDA flow, particularly for RTL tasks, promises to enhance modern chip design and accelerate time-to-market. 
% To enhance the current EDA tools for complex modern chip design and expedite time-to-market, integration of Artificial Intelligence (AI) techniques into the EDA flow, particularly for RTL-related tasks, has emerged as a promising avenue. 
For instance,\cite{allam2024RTL-Repo} employs large language models (LLMs) to understand and describe RTL designs in natural language to assist engineers.\cite{cui2024OriGen,thakur2023benchmarking} focus on generating RTL code from the design specification. Clearly, the effectiveness of deep learning models is heavily contingent upon the quality of the training data~\cite{chang2024dataisall}. However, we observe that existing RTL datasets suffer from notable limitations, hindering the application of AI-based solutions in practical RTL modeling and verification.

% Modern chip design is a multifaceted endeavor characterized by its intricacy, involving a diverse array of sub-modules, alongside numerous stages within the Electronic Design Automation (EDA) flow. The process of designing typically entails collaboration across many departments in a semiconductor company to translate the design specification into Register Transfer Level (RTL) code, netlist, floorplanning and layout. 

% To enhance the existing EDA tools and speed time-to-market, integration of Artificial Intelligence (AI) technique in EDA has emerged as an attractive direction. For instance, DeepGate Family~\cite{li2022deepgate, shi2023deepgate2, shi2024deepgate3} are trained on a large number of circuit netlists and employed into design for test~\cite{shi2022deeptpi} and power analysis~\cite{khan2024deepseq}. Besides, ~\cite{allam2024RTL-Repo, cui2024OriGen, thakur2023benchmarking} learn web-scale code and benefit hardware code generation. Clearly, the effectiveness of deep learning models is heavily contingent upon the quality of the training data~\cite{chang2024data}. However, we observe that the existing circuit datasets in EDA field suffer from notable limitations.

% Existing dataset: 1. for a single task (EPFL, ISCAS89, ITC99, NYU) 2. less types (CircuitNet), not suitable for AI4EDA. 
One prominent limitation is the narrow scope of most datasets. Since the semiconductor ecosystem is far inferior to software openness, the accessible designs are limited. Many datasets either focus on a limited set of circuit types, often centering around a single design such as processors~\cite{chai2022circuitnet, jiang2024circuitnet2}), or indiscriminately collect all Verilog files without accounting for their correctness and distribution~\cite{thakur2023benchmarking, wu2024edacorpus}. This restricted diversity hampers the generalizability and effectiveness of these datasets.

% Another critical limitation lies in the task-specific nature of existing datasets. Understanding the interconnection across various stages of the EDA process is crucial for optimizing AI-based EDA approaches. Minor alterations in the front-end design stages can significantly impact back-end processes and the final Power, Performance, and Area (PPA) metrics. However, the existing datasets~\cite{chowdhury2021openabc, lu2024rtllm} are typically designed for singular tasks, overlooking the essence of chip design flow as a multimodal circuit data transformation. 
% Another critical limitation is the oversight of the circuit implementation based on RTL design. For example, RTLLM~\cite{lu2024rtllm} gathers RTL specifications and manually verifies the correctness of the corresponding RTL code, while\cite{chang2024dataisall} ensures the functionality when constructing and augmenting datasets. However, these datasets overlook the correlation between RTL code and the circuit implementations (e.g. netlist, layouts) in the following stages of EDA flow. Therefore, training an RTL model with the existing datasets that both guarantees design functionality correctness and optimizes RTL code to achieve better Power, Performance and Area (PPA) remains challenging. 
Another critical limitation is the lack of focus on circuit implementation in RTL-based datasets. For instance, RTLLM~\cite{lu2024rtllm} gathers RTL specifications and verifies the functional correctness of RTL code, while\cite{chang2024dataisall} ensures functionality when constructing and augmenting datasets. RTL-Repo\cite{allam2024RTL-Repo} collects 1000+ RTL design projects and make each sample in the dataset contains the context of the entire repository to train LLM for RTL code understanding. However, these works do not consider or include circuit-level information, such as netlists and layouts, which are crucial in later stages of the EDA flow. As a result, these datasets overlook the correlation between RTL code and its corresponding circuit implementations. Consequently, training RTL models with the existing datasets that both guarantee functional correctness and optimize Power, Performance, and Area (PPA) remains a significant challenge.

% Our dataset: Collect based on the real-world chip types, Collect repo-level, synthesis flow
To overcome these limitations, we present a multimodal and comprehensive
repository-level circuit dataset for deep learning in EDA. Our dataset is meticulously curated according to the scope of the real-world chip and collect more than 4,000 circuit design projects from various data sources. Unlike previous datasets that provide verilog files only\cite{chang2024dataisall,thakur2023benchmarking}, each data point in our dataset is structured as a complete repository and also splitted into files, modules and blocks for various scenarios. This multi-tiered structure enables training models at different scales, making it suitable for a variety of LLMs and models. Moreover, the repository-level data allows logic synthesis and physical design flows to convert RTL code into various circuit modalities, including Control/Data Flow Graph (CDFG) of RTL code, And-Invertor Graph (AIG), post-mapping netlist, floorplaning and layout. Therefore, we can build the interactions between RTL designs and circuit implementations. 

% We label these circuits for unimodal RTL tasks and cross-modal PPA prediction. 
Furthermore, we meticulously label the circuits for unimodal RTL tasks and cross-modal PPA prediction to enhance accessibility of our dataset. We propose a Chain of Thought (CoT)\cite{wei2022chainofthought} detailed annotation method to generate descriptions and comments for each of the four levels, namely, repo-level, file-level, module-level and block-level.  By using GPT-4\cite{achiam2023gpt} and Claude\cite{claude}, we leverage annotations from higher levels to assist in annotating lower levels. Additionally, we generate question-answer pairs to help describe the functionality and key features of each code segment, enabling better training data for LLMs. Moreover, we provide the corresponding logic synthesis results for the cross-stage PPA prediction on RTL code. 

% With this dataset, we can construct a variety of pre-training tasks, evaluation tasks, and benchmarks, enabling the study of LLM performance on RTL code understanding and completion. Our experiments include tasks that assess LLMs' abilities in RTL code comprehension, generation and completion, we trained various models, including CodeLLama\cite{codellama}, CodeT5+\cite{codet5}, CodeGen\cite{codegen}, and DeepSeek\cite{liu2024deepseekv2,zhu2024deepseekcoderv2}, across different scales ranging from 220M to 16B on our dataset. The results demonstrate two key findings:
% 1)
% Every large model fine-tuned on our dataset significantly outperforms its original, non-fine-tuned counterpart across all metrics, highlighting the effectiveness of our data.
% 2) 
% LLMs of different scales, such as the 220M CodeT5, 7B and 16B models, show substantial improvements after fine-tuning, reflecting the adaptability and generalization capabilities of our dataset across varying model sizes. Moreover, to prove the contribution of our dataset on the EDA tasks, we utilize learning-based PPA prediction models, the results show that accurate predicting PPA on early stage still poses a question, providing valuable insights for future research in RTL and hardware design automation. 
With this dataset, we create a range of pre-training tasks, evaluation tasks, and benchmarks. One experiment focuses on RTL code understanding and generation with LLMs. We trained models including CodeLLama\cite{codellama}, CodeT5+\cite{codet5}, CodeGen\cite{codegen}, and DeepSeek\cite{liu2024deepseekv2, zhu2024deepseekcoderv2} at scales ranging from 220M to 16B parameters on our dataset. Another experiment evaluates PPA using learning-based prediction models\cite{xu2022sns,sengupta2022good,fang2023masterrtl}, trained to assess area, power, and delay. The results reveal three key findings:
1) Fine-tuning LLMs on our dataset leads to significant performance improvements across all metrics, demonstrating the effectiveness of our dataset.
2) LLMs of different scales, such as CodeT5 220M, 7B, and 16B, show substantial gains after fine-tuning, indicating the adaptability of our dataset across varying model sizes.
3) PPA prediction results highlight the ongoing challenge of accurate early-stage PPA forecasting, providing valuable insights for future RTL and hardware design automation research.
And the pipeline and framework are illustrated in Figure \ref{fig:overview}.

The main contributions of our work are summarized as follows:
\begin{itemize}
  % real-world, ->multimodal (all tasks in EDA) -> annotations
    \item We propose a holistic dataset including over 4,000 repository-level RTL projects, covering chip-designs, IP-designs, module-designs designs, and incorporating a diverse range of functional and algorithmic keywords.

    \item Our dataset is organized into four levels, namely, repository, file, module, and block levels, allowing models to be trained at different scales and enabling broader applications in EDA tasks such as synthesis, netlist generation, PPA analysis, and layout design.

    \item We propose a Chain of Thought (CoT) annotation method using GPT-4 and Claude to generate detailed comments, descriptions, and question-answer pairs, improving training data for LLMs on the tasks of RTL code understanding and generation.
    
    \item We create pre-training and evaluation benchmarks for LLMs on RTL tasks such as code understanding, completion, and neural network-based PPA prediction, demonstrating the effectiveness, adaptability and generalization capabilities of our dataset for both RTL code comprehension and EDA tasks.
\end{itemize}

\section{Related Work} \label{Sec:Related}
\subsection{Previous Dataset in EDA}
AI-based methodologies excel in addressing classification, prediction, and optimization tasks, making them well-suited for chip design. Over the past decade, the integration of AI into Electronic Design Automation (EDA) has emerged as an attractive direction in the fields of chip design and semiconductor industry, which is well surveyed in~\cite{huang2021machine, chen2024dawn}. 

The cornerstone of training AI models lies in the quality of the dataset utilized. In the domain of EDA, existing datasets can be broadly categorized into two classes. Firstly, there are unimodal datasets or benchmarks such as ISCAS'89~\cite{ISCAS89}, ITC'99~\cite{ITC99}, IWLS'05~\cite{IWLS05} and EPFL~\cite{EPFLBenchmarks} benchmarks, which offer open-source circuit netlists primarily for front-end applications like logic synthesis and design for test. \cite{shrestha2024edaschema} presents a dataset of physical designs generated from the IWLS'05 benchmark circuit suite, utilizing the open-source 130nm Process Design Kit (PDK) by Skywater and the OpenROAD toolkit \cite{ajayi2019openroad}. 
% Additionally,~\cite{thakur2023benchmarking} collects Verilog code sourced from the Internet and ~\cite{wu2024edacorpus} assembles a dataset in a natural language format containing question-answer pairs. 
Despite these efforts, these datasets still lack in providing comprehensive insights into cross-stage EDA tasks, such as optimizing circuit designs at the front-end for improved PPA metrics during the back-end stage.

Secondly, other multimodal circuit datasets are generated using EDA tools. For instance, CircuitNet~\cite{chai2022circuitnet, jiang2024circuitnet2} creates a dataset through logic synthesis and physical design, where the circuit designs are synthesized into gate-level netlists and transformed into layouts using commercial EDA tools. \cite{jiang2024circuitnet2} expand this work by providing data for million-gate designs such as CPUs, GPUs, and AI chips, and using the 14nm FinFET technology node to capture the increased complexity of manufacturing and modeling. Nonetheless, the existing datasets encompass a limited range of circuit designs, thereby constraining the generalizability of subsequent model training efforts. 

% \cite{zhong2023llm4eda} explore the wide-ranging applications of LLMs in code analysis, including bug detection and fixing, code summarization, and security checking. They also introduce the use of LLMs in logic synthesis, physical design, and the generation of HDL and scripts. \cite{wu2024edacorpus} release a prompt-script dataset to train LLMs for script generation in physical design tasks using OpenROAD and propose a question-answer dataset to train LLMs on answering questions related to physical design methods with OpenROAD.  \cite{chai2022circuitnet} creates a dataset through logic synthesis and physical design, where RISC-V designs are synthesized into gate-level netlists and transformed into layouts using Synopsys and Cadence tools, with diverse settings to improve dataset variability. \cite{jiang2024circuitnet2} expand this work by providing data for million-gate designs such as CPUs, GPUs, and AI chips, and using the 14nm FinFET technology node to capture the increased complexity of manufacturing and modeling. This dataset supports multi-model prediction tasks, including timing, routability, and IR-drop, for advanced technology nodes, covering a broader range of design objectives.

\subsection{RTL-stage Understanding, Completion, and Generation}
\cite{thakur2023benchmarking} collects approximately 50,000 open-source Verilog code samples and fine-tunes five pre-trained LLMs, with model sizes ranging from 345 million to 16 billion parameters. \cite{liu2023verilogeval} introduces a comprehensive evaluation dataset consisting of 156 problems sourced from HDLBits and developed a benchmarking framework to automatically test the functional correctness of Verilog code completions. Similarly, \cite{thakur2024verigen} fine-tunes existing LLMs on Verilog datasets collected from GitHub and textbooks, evaluating the functional correctness of the generated code using a custom test suite. \cite{chang2024dataisall} designs a data augmentation framework for training Chip Design LLMs, enabling them to generate Verilog code, EDA scripts, and coordinate EDA workflows based on natural language design descriptions. They also benchmark their approach by fine-tuning Llama 2 models with 7 billion and 13 billion parameters. Lastly, \cite{zhang2024mg} presents the MG-Verilog dataset, an open-source dataset that meets essential criteria for high-quality hardware data, facilitating the effective use of LLMs in hardware design. However, these datasets focus solely on file-level RTL code, neglecting the comprehensive information contained within entire RTL project designs, modules, and code blocks. As a result, they fail to provide the multimodal data, such as netlists and PPA metrics, that can be obtained through synthesis. 
Moreover, existing studies primarily focus on RTL code generation and completion, often overlooking the critical aspects of annotations, comments, and descriptions for RTL code. Hence, there has been little progress in advancing RTL code understanding tasks, particularly for LLMs.

% Furthermore, existing works primarily concentrate on RTL code generation tasks, often \textcolor{red}{comment: cannot understand this sentence: }overlooking the essential aspects of annotation and the construction of understanding tasks for LLM.

% \subsubsection{Cross-stage PPA Prediction}
\section{Dataset} \label{Sec:Dataset}
\subsection{Data Preparation}

To maximize the utility of our dataset and enable training and evaluation for LLMs, it is essential to clearly distinguish between chip-level, IP-level, and module-level RTL designs. The repositories of different levels differ in abstraction, scope, and their roles in hardware design automation. 
In detail, chip-level design encompasses the entire System-on-Chip (SoC) and processors, focusing on system-wide integration, interconnects, and global constraints like power, timing, and area. 
IP-level design addresses modular and reusable components, such as cores or memory controllers , with a focus on standalone functionality and scalability. 
Module-level design targets subcomponents within IPs, such as arithmetic units or fundamental operators, emphasizing detailed functionality and local optimization. These levels differ in scope, with chip-level representing the holistic system, IP-level the building blocks, and module-level the internal details, enabling hierarchical abstraction and efficient hardware design.

\begin{figure}[ht]
    \centering
    \includegraphics[width=\linewidth]{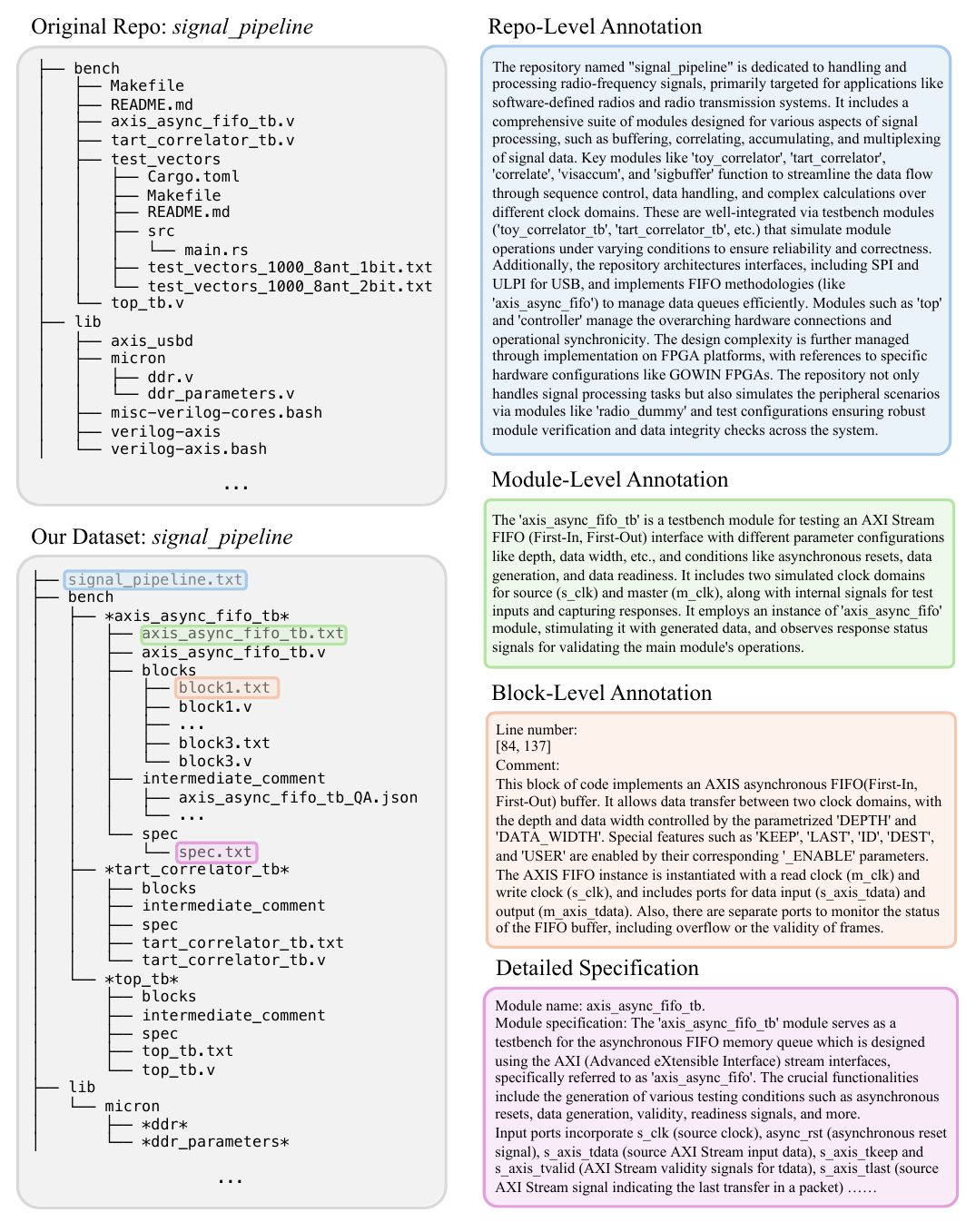}
    \caption{Illustration of the dataset repository structure with multi-level annotations}
    \label{fig:overview_data}
\end{figure}

Our dataset distinguishes itself by focusing on chip-level, IP-level, module-level RTL designs. We collect a list of 222 keywords representing these levels from sources like Alldatasheet~\cite{alldatasheet} and other relevant websites. Examples include chip-level designs such as voice processors, audio processors, and video processors; IP-level designs like DMA, PCI, and true random number generators; and module-level designs including 4-bit binary full adders, arithmetic logic units, and multiplier-accumulators. We collect over 4,000 repository-level RTL projects, which encompass 140,000 RTL files across 77 functional categories, as shown in Table \ref{tab:dataset_levels}. Moreover, the detailed data cases are illustrated in Figure \ref{fig:overview_data}. 

To construct the RTL-language dataset, we organize the data into four distinct levels: repository, file, module, and block. We employ a Chain of Thought (CoT) approach for RTL code annotation, leveraging GPT-4~\cite{achiam2023gpt} and Claude~\cite{claude} to generate detailed comments, descriptions, and question-answer pairs. This methodology enhances the training data for large language models (LLMs) in RTL code understanding and generation, with further details provided in Section \ref{sec:RTL_annotation}. Additionally, to develop the multimodal dataset, we synthesize the RTL projects to obtain netlists, power, performance, and area (PPA) metrics, as well as layout designs. Further details will be discussed in Section \ref{sec:RTL_Repo_synthesize}.

\begin{table}[ht]
\centering
\caption{Summary of data across different levels in DeepCircuitX. The table presents the number of function categories, repositories, and RTL files at the Chip, IP, Module designs.}
\vspace{1em}
\resizebox{\linewidth}{!}{
\begin{tabular}{l|ccc}
\toprule

\textbf{Level} & \textbf{Function Categories} & \textbf{Repo Number} & \textbf{RTL File Number} \\ \midrule
Chip Level     & 17   & 1002   & 54650  \\
IP Level       & 3    & 1410  & 92467 \\ 
Module Level   & 57   & 2383  & 38692  \\  \bottomrule
\end{tabular}}
% \caption{Summary of data across different levels}

\label{tab:dataset_levels}
\end{table}

\subsection{Language-RTL Code Dataset Construction}
\label{sec:RTL_annotation}
\subsubsection{Chain of Thought (CoT) for RTL Code Annotation}
The Chain of Thought (CoT) reasoning method, commonly used in deep learning and natural language processing, simulates human reasoning by guiding models through structured steps. We adopt this approach to create detailed annotations for Verilog code, ensuring a comprehensive understanding at different levels of the RTL design. The annotation process is carried out at three distinct levels: \textbf{\textit{module-level}}, \textbf{\textit{block-level}}, and \textbf{\textit{repository-level}}. 
% The detailed pipeline is shown in Figure \ref{fig:cot} in the Appendix.

% Chain of Thought (CoT) is a reasoning method that has been widely applied in deep learning and natural language processing in recent years, aiming to simulate the process of human thought by gradually guiding the reasoning steps. To generate accurate descriptions for Verilog code, we apply the CoT approach to create annotations. The detailed pipeline is illustrated in Figure \ref{fig:cot}.
\paragraph{Module-Level Annotations} To annotate RTL modules, we utilize a multi-round question-answering approach to extract key information from the Verilog code, focusing on aspects such as the module's name, input/output ports, and internal signals. This process begins with structured questions designed to clarify both what the module does (What) and how it achieves this functionality (How). The responses provide the foundational details about the module's components and behavior. Using this information, a detailed specification is created that summarizes the module’s name, its purpose, the roles of the input and output ports, and an explanation of the internal signals. Additionally, the specification includes a breakdown of the functional blocks within the module. Finally, with this specification and the original code, ChatGPT generates a concise yet comprehensive module-level annotation, encapsulating both the functionality and the implementation details.

\paragraph{Block-Level Annotations} Block-level annotations offer detailed insights into the functionality of specific sections within an RTL module. Given the complexity of Verilog code, which often includes nested structures, segmenting it using regular expressions alone can be challenging. To address this, we leverage GPT-4 for handling these intricacies. The code is divided into distinct functional blocks, including constructs such as \texttt{always}, \texttt{initial}, \texttt{task}, \texttt{function}, \texttt{generate}, \texttt{assign}, and \texttt{final} blocks. While regular expressions are employed to improve segmentation accuracy for straightforward patterns, GPT-4 is crucial for managing more complex structures. Once the blocks are identified, each is annotated to describe both its purpose and how it functions, ensuring that the role of each block within the module is clearly defined, thereby enhancing the overall understanding of the module's functionality.

\paragraph{Repo-Level Annotations} At the repository level, we aim to provide an overarching understanding of the entire RTL project. This includes information about the structure, purpose, and interconnections of various modules and files within the repository. To achieve this, we gather the file structure information and combine it with module-level annotations to form a complete picture of the project. GPT-4 is then used to summarize the contents of the repository, producing a top-down view of how individual files and modules work together. This approach allows the repo-level annotation to reflect both functional and design-level details, capturing the broader goals of the RTL project.

% \begin{figure}[!tb]
%     \centering
%     \includegraphics[width=\linewidth]{fig/iclr_Dataset_cot.pdf}
%     \caption{Detailed procedures in our proposed method: CoT code-annotation.}
%     \label{fig:cot}
% \end{figure}

\begin{table}[ht]
\centering
\caption{Summary of annotated RTL categories at the module, block, and repository levels in DeepCircuitX. The table highlights the number of annotated modules, blocks, and repositories for each RTL category, including Chip, IP, and Module.}
\vspace{1em}
\begin{tabular}{l|ccc}
\toprule
\textbf{RTL categories} & \textbf{Module-Level} & \textbf{Block-Level } & \textbf{Repo-Level }  \\ \midrule
Chip  & 7587    & 36955    & 684       \\
IP  & 12863    & 20101    & 183       \\ 
Module  & 28901    & -    & 1389       \\ \bottomrule
\end{tabular}
% \caption{Summary of annotations across different levels}

\label{tab:dataset_repo_levels}
\end{table}

\subsubsection{Dataset for RTL Code Understanding, Completion and Generation}

In addition to annotations, we constructed a dataset that supports three distinct tasks related to RTL code: \textbf{\textit{understanding}}, \textbf{\textit{completion}}, and \textbf{\textit{generation}}. Table \ref{tab:tasks_categories} shows the data counts across RTL categories (IP, Module, and Chip) for each task.

\begin{table}[htbp]
    \centering
    \caption{Data counts for code generation, code completion, and comment generation tasks across different categories.}
    \vspace{1em}
    \begin{tabular}{l|rrrrr}
        \toprule
        \textbf{Tasks Dataset} & \textbf{IP} & \textbf{Module}  & \textbf{Chip} & \textbf{All} \\
        \midrule
        RTL Understanding & 6386 & 14499 & 5270 & 26155 \\
        RTL Completion & 6178 & 14131 & 5134 & 25443 \\
        RTL Generation & 6479 & 16511 &  5343 & 28333 \\
        \bottomrule
    \end{tabular}

    \label{tab:tasks_categories}
\end{table}

% \textbf{RTL Code Understanding} 

% In the RTL code understanding task, the goal is to evaluate the model's ability to comprehend and generate a meaningful description of a given RTL code. In this task, we provide the RTL code as input, and the target output is a detailed, concise description of the functionality and structure of the code. The description includes information about the module’s purpose, its input/output signals, internal logic, and the overall behavior. This task helps assess how well models can interpret RTL code and produce human-readable explanations, which is critical for documentation and code analysis purposes.
\paragraph{RTL Code Understanding} This task evaluates the model’s ability to interpret and describe RTL code. Given a module’s RTL code as input, the model generates a detailed, concise description, covering key aspects such as the module’s purpose, input/output signals, internal logic, and overall behavior. This task is crucial for assessing the model’s ability to generate human-readable explanations for code analysis and documentation.

% \textbf{RTL Code Completion} 

% The RTL code completion task focuses on the model’s ability to generate the missing parts of RTL code based on a partial input. We provide a description of the code, along with the module header, which includes the input/output ports or some parameters. The target is for the model to complete the rest of the code, implementing the internal logic, control structures, and signal definitions. This task evaluates the model’s capability to infer and generate the missing code based on context, similar to autocompletion in text editors, and is useful for enhancing productivity in RTL design.
\paragraph{RTL Code Completion} In this task, the model is provided with a partial RTL code (typically the module header with input/output ports and parameters). The goal is for the model to complete the code by generating the missing internal logic, control structures, and signal definitions. This task mirrors autocompletion functionality found in modern code editors and evaluates the model’s ability to infer and generate code from context.

% \textbf{RTL Code Generation} 

% In the RTL code generation task, the model is tasked with generating a complete implementation of RTL code based on a given description and specific details about the I/O and parameters. The description provides a high-level overview of the module’s intended functionality, while additional information, such as the input/output ports and their configurations, guides the code generation process. The goal is to produce a fully functional Verilog module that adheres to the provided specifications. This task evaluates the model’s ability to transform high-level design requirements into a precise RTL implementation, which is essential for automating the hardware design process.
\paragraph{RTL Code Generation} 
In the RTL code generation task, the model is tasked with producing a full implementation of RTL code based on a high-level description and specified input and output parameters. The goal is to generate a fully functional Verilog module that adheres to the provided specifications. This task assesses the model’s ability to translate design requirements into precise RTL implementations, which is critical for automating the hardware design process.

\subsection{Multimodal Transformation of RTL Code}
\subsubsection{Graph-based Code Representation}
Abstract Syntax Tree (AST) serves as a tree-based representation to model the structure of programming code, extensively used in syntax analysis and compilation. In the context of Verilog code, each node within the AST denotes a variable or operator, and the edges formulate the relations between these nodes. Additionally, Control/Data Flow Graph (CDFG)~\cite{orailoglu1986flow} is a another graph-based code representation to visualize how data moves between registers and how control signals dictate the operation of the design. CDFG proves advantages for modeling and verifying the functionality of RTL designs~\cite{coussy2009introduction, vasudevan2021learning}. We generate the AST and CDFG of RTL designs with open-source tools Yosys~\cite{wolf2016yosys} and customized Pyverilog~\cite{takamaeda2015pyverilog}.

\subsubsection{Circuit Netlist Synthesis}
\label{sec:RTL_Repo_synthesize}
Our dataset contains RTL repositories with module invocations and complete Verilog files, enabling us to derive the circuit netlists via logic synthesis process. For each RTL repository, we employ the commerical tool Synopsys Design Compiler 2019.12 to transform HDL code into netlists. The RTL designs are mapped into several open-source technology libraries, including GlobalFoundries 180nm, skywater 130nm, ihp-sg 130nm, nangate 45nm and asap 7nm. Each RTL file is synthesized using both the compile and \textit{compile\_ultra} commands. By toggling the set max area 0 option, we obtain both the default netlist and a netlist optimized for maximum area constraints. We assess the internal, transition, and leakage power of each mapped netlist using PrimeTime (Synopsys, 2023.12). The critical path delay is reported by the Design Compiler. 

\begin{figure}
    \centering
    \includegraphics[width=\linewidth]{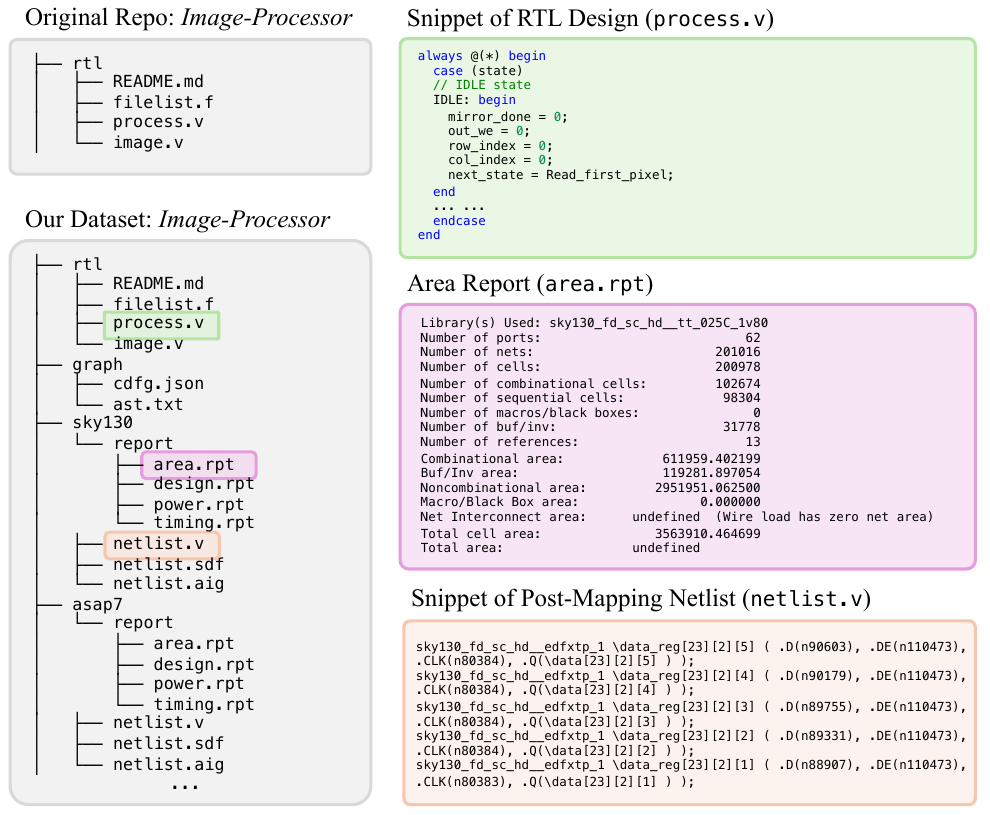}
    \caption{Illustration of the dataset repository structure with Circuit Transformation}
    \label{fig:multimodal}
\end{figure}

Finally, we store the post-mapping netlist into both verilog format (.v files) and Standard Delay Format (SDF)~\cite{sagdeo1998standard} (.sdf files). Moreover, we record the logic synthesis reports (.rpt files) with maximum path delay, area and dynamic power for the following experiments. We also convert the post-mapping netlist into And-Inverter Graph (AIG) format, a prevalent and widely-adopted  common format for netlist learning~\cite{li2022deepgate, shi2023deepgate2, shi2024deepgate3, deng2024less}, using abc~\cite{brayton2010abc} tool for further investigation. The multi-modal data cases are illustrated in Figure \ref{fig:multimodal}.

\section{Benchmark and Experiments} \label{Sec:TradApp}
% \section{Benchmark and Experiments}
\subsection{overview}
To evaluate our multi-modal dataset DeepCircuitX and establish benchmarks, we design experiments of human evaluation in Section \ref{sec:humaneval}, RTL code tasks in Section \ref{sec:understanding} and Section \ref{sec:completion}, and PPA prediction experiments in Section \ref{sec:PPA}.

Specially, in Section \ref{sec:understanding} and Section \ref{sec:completion}, we establish benchmarks for LLMs tailored to RTL tasks, utilizing open-source models such as CodeLlama~\cite{codellama}, CodeT5+~\cite{codet5}, deepseek-coder~\cite{guo2024deepseekcoder} and CodeGen~\cite{codegen}. Our annotated data is employed to fine-tune these models for RTL code understanding, completion, and generation tasks. Here is an concise iintroduction of the based-LLMs we select:
\begin{itemize}
    \item CodeLlama is fine-tuned Llama on a diverse corpus of code from various programming languages, enabling it to understand syntax, semantics, and the context of code snippets. 
    \item CodeT5+ is an enhanced version of the original CodeT5 model, specifically optimized for code understanding and generation tasks. Building on the strengths of its predecessor.
    \item CodeGen is trained on GPT-3~\cite{floridi2020gpt3} by a diverse range of programming languages and employs advanced techniques to understand and generate code efficiently. 
    \item DeepSeek-V2-lite~\cite{liu2024deepseekv2} has broader applicability on general tasks of natural language and DeepSeek-Coder-V2-lite~\cite{zhu2024deepseekcoderv2} focuses on code-related tasks, offering fine-tuned capabilities for high-quality code generation and synthesis.
\end{itemize}

\subsection{Human evaluation}
\label{sec:humaneval}
To evaluate the effectiveness of our generating comments approach and the quality of the annotations, we conduct a series of evaluation criteria and metrics involving human reviews by independent experienced engineers. 
We employ the following metrics: Accuracy, Completeness and Understandable Clarity. The detailed grading criteria is introduced in the Table \ref{tab:grading_criteria}.

\begin{table}[htbp]
\centering
    \caption{Grading Criteria for Annotation Quality.}
\label{tab:grading_criteria}
\resizebox{\linewidth}{!}{
\begin{tabular}{c|c|c|c}
\hline
\textbf{Grade} & \textbf{Accuracy}         & \textbf{Completeness}   & \textbf{Understandable Clarity} \\ \hline
4              & Completely Accurate       & Fully Complete          & Clear and Concise               \\ \hline
3              & Partially Accurate        & Mostly Complete         & Relatively Clear                \\ \hline
2              & Not Quite Accurate        & Not Quite Complete      & Vague                           \\ \hline
1              & Completely Incorrect      & Incomplete              & Inaccurate         \\ \hline
\end{tabular}}
\end{table}

Engineers are tasked with analyzing both the RTL code and the accompanying annotations based on specific criteria, ensuring a thorough evaluation of the provided information. Subsequently, they assign a grade to each metric on a scale from 1 to 4, with 4 indicating the highest quality and 1 indicating the lowest. To ensure fairness, each generated text is reviewed by 5 individuals, and the average score is recorded in Table \ref{tab:human_eval}. We observe that the code annotations in DeepCircuitX exhibit high quality, with all metrics scoring above 3.5 out of 4.

\begin{table}[ht]
\centering
\caption{Human evaluation grading of repo-level annotation and module-level annotation, we use the metrics of accuracy, completeness, understandable clarity to evaluate the quality of our annoation.}
\vspace{1em}
\begin{tabular}{lcc}
\toprule
\textbf{Metrics} & \textbf{Repo Annotation} & \textbf{Module Annotaion} \\ 
\midrule
Accuracy &3.74/4		&3.5/4	\\
Completeness &3.79/4   &3.78/4\\
Understandable Clarity &3.84/4   &3.76/4\\
\bottomrule
\end{tabular}

\label{tab:human_eval}
\end{table}

\subsection{RTL Code Understanding}
\label{sec:understanding}

% \subsubsection{Overview}
% For the task of RTL code understanding, we select LLama2 (7b), CodeLLama (7b), CodeT5+ (220m), CodeT5+ (6b), CodeGen2 (1b), CodeGen2.5 (7b), DeepSeek-Coder-V2-lite (16b), DeepSeek-V2-lite (16b) as our based LLM. And we use our dataset to inst

\subsubsection{Evaluation Metrics}

\textbf{BLEU} measures n-gram overlap (1 to 4) between generated text and reference translations, incorporating a brevity penalty.
\textbf{METEOR} accounts for synonymy and stemming, calculating a harmonic mean of precision and recall. 
\textbf{ROUGE} focuses on summarization, with variants like ROUGE-N for n-gram overlap and ROUGE-L for the longest common subsequence, emphasizing recall to assess content relevance.

\subsubsection{Analysis}

The experimental results, shown in Table~\ref{tab:exper_understanding}, demonstrate the performance of various base Large Language Models (LLMs) and fine-tuned LLMs on our RTL code understanding benchmark, using evaluation metrics BLEU-4, METEOR, ROUGE-1, ROUGE-2, and ROUGE-L, which offer valuable insights into the quality of generated RTL code in terms of surface-level linguistic similarity.

Initially, the original versions of the LLMs, such as CodeLLama, CodeT5+, CodeGen2, and DeepSeek, exhibit relatively low performance across most metrics. For example, CodeLLama (original) achieves a BLEU-4 score of 0.0828, while CodeGen2.5 (original) shows moderate improvement with a BLEU-4 score of 0.1060. Among the original models, DeepSeek-Coder-V2-lite (original) stands out, significantly outperforming others with a BLEU-4 score of 2.2387 and a ROUGE-1 score of 30.3208, indicating its strong baseline performance even before fine-tuning.

After fine-tuning on our dataset, every large model demonstrates significantly better performance across BLEU-4, METEOR, ROUGE-1, ROUGE-2, and ROUGE-L metrics compared to their original, non-fine-tuned counterparts. This highlights the effectiveness of our dataset. Moreover, models of various sizes, such as the 220M CodeT5, as well as larger 7B and 16B models, all show substantial improvements after fine-tuning. This indicates that our dataset is well-suited for models of different scales, providing strong adaptability and generalization.
% However, fine-tuning the models on our training dataset leads to substantial performance improvements across the board. CodeLLama (7b), for instance, sees a dramatic rise in BLEU-4 and similar gains in ROUGE-1 and METEOR scores. CodeT5+ (220m-bimodal) demonstrates one of the most notable improvements, achieving a high BLEU-4 score and ROUGE-1 score, far surpassing its original version. Similarly, CodeGen2.5 (7b) shows significant enhancements in BLEU-4, METEOR, and all ROUGE metrics, highlighting the effectiveness of fine-tuning.

% In the larger models, DeepSeek-V2-lite (16b) and DeepSeek-Coder-V2-lite (16b) also exhibit strong performance after fine-tuning, along with high ROUGE scores across all categories. These results suggest that LLMs benefit considerably from fine-tuning, particularly present the effectiveness of our RTL code understanding dataset.

Overall, the fine-tuned LLMs significantly outperform their original counterparts, indicating the importance of fine-tuning on domain-specific datasets like ours.

\begin{table}[ht]
\centering
\caption{The results of our base-LLMs and fine-tuned LLMs by our training dataset on our RTL code understanding benchmark.}
\vspace{1em}
\resizebox{\linewidth}{!}{
\begin{tabular}{lccccc}
\toprule
\textbf{Based LLM} & \textbf{BLEU-4} & \textbf{METEOR} & \textbf{ROUGE-1} & \textbf{ROUGE-2} & \textbf{ROUGE-L} \\ 
\midrule
CodeLLama (original) & 0.0828 & 5.9414& 11.0046& 0.3139 & 0.2581       \\ 
CodeT5+ 220m-bimoal (original) & 0.1410    & 4.3277    & 10.9925 & 0.5076 & 9.3043      \\ 

% CodeT5+ 6b (original)        \\ 
CodeGen2 (original)  & 0.1082 &3.7890  &8.0311 &0.1173    & 7.2161 \\ 
CodeGen2.5 (original)  & 0.1060 & 4.8271 & 9.4001 & 0.3698 & 8.5856         \\ 
DeepSeek-Coder-V2-lite (original)&2.2387 &24.3311 &30.3208 &6.3163 & 27.4282        \\ 
DeepSeek-V2-lite (original) &0.2311 &3.3951 &7.3720 &0.2646 &6.3360     \\ 
\midrule
CodeLLama (7b) & 0.8619 & 18.2621 & 25.0461 & 6.4493 & 22.8182      \\ 
CodeT5+ (220m-bimodal) & 4.9067    & 23.5043   & 34.8671 & 9.9023 & 32.1642      \\ 
% CodeT5+ (6b)        \\ 
CodeGen2 (1b)  & 7.7605 &27.8049 & 37.2150 & 13.1385 & 34.0647        \\ 
CodeGen2.5 (7b) & 13.6858 &34.7494 &43.5244 &18.5249 &40.2223      \\ 
DeepSeek-Coder-V2-lite (16b) &11.9180 &33.5011 &41.8527 &17.2014 &38.0473      \\ 
DeepSeek-V2-lite (16b) & \textbf{13.6972} & \textbf{39.5962} & \textbf{43.3732} & \textbf{19.0589} & \textbf{39.5962}      \\ 
\bottomrule
\end{tabular}}

\label{tab:exper_understanding}

\end{table}

\subsection{RTL Code Completion and Generation}
\label{sec:completion}

% \subsubsection{Overview}
% For the task of RTL code completion and generation, we select LLama2 (7b), CodeLLama (7b), CodeT5+ (220m), CodeT5+ (6b), CodeGen2 (1b), CodeGen2.5 (7b), DeepSeek-Coder-V2-lite (16b), DeepSeek-V2-lite (16b) as our based LLM. And we use our dataset to inst

\subsubsection{Evaluation Metrics}
In the realm of RTL code completion and generation, the evaluation of model performance is critical to advancing intelligent programming tools. The Pass@k metric serves as a pivotal measure in this domain, quantifying the accuracy of code generation models by assessing their ability to produce valid solutions within the top-k predictions. Specifically, Pass@k evaluates whether the correct code snippet appears among the model's top k outputs, thereby providing insights into both the effectiveness and reliability of the model's predictions.

\subsubsection{Analysis}

Table \ref{tab:exper_gen_com} compares the performance of both original and fine-tuned LLMs on RTL code completion and generation tasks, focusing on Pass@1 and Pass@5 on two evaluation benchmarks, RTLLM\cite{lu2024rtllm} and VerilogEval\cite{liu2023verilogeval}. 
We choose the baseline LLMs as original versions of CodeLlama, CodeT5+, CodeGen2, and CodeGen2.5 exhibit negligible performance, with most Pass@K scores at 0\% or near 0\%.

Notably, every model fine-tuned with our dataset significantly outperforms its original, non-fine-tuned counterpart, demonstrating the effectiveness of our data. Additionally, for models of different scales, such as the 220M CodeT5 and 7B models, the results after fine-tuning show substantial improvements. This highlights the adaptability and generalization capability of our dataset across various model sizes. Moreover, we include CodeV (QW-7B) \cite{zhao2024codev} as an additional baseline, which achieves 14.80\% Pass@1 and Pass@5 on RTLLM, and 4.5\% on VerilogEval. Although CodeV has undergone prior fine-tuning for general-purpose code generation, its performance remains lower than our fine-tuned CodeGen2.5 (7B). 

These findings highlight the effectiveness of our dataset in enhancing LLMs’ capability to generate syntactically and functionally accurate RTL code. Future work could explore further model scaling and more advanced training techniques to push performance even higher.

\begin{table}[htbp]
\centering
\caption{Pass@K results for RTL code completion and generation across various LLMs. The result with ∗ are evaluated using our custom implementation to align with our experimental settings.}
\vspace{1em}
\setlength\tabcolsep{1.0pt}
\begin{tabular}{c|cc|cc}
\toprule
\multirow{2}{*}{Based LLM} & \multicolumn{2}{c|}{Pass@1}   & \multicolumn{2}{c}{Pass@5} \\
                           & RTLLM\cite{lu2024rtllm}       & VerilogEval\cite{liu2023verilogeval}    & RTLLM       & VerilogEval    \\ \midrule
CodeLLama (original)              & 0           & 0.64\%        & 0           & 0 \\
CodeT5+ (original)   & 0          & 0             & 0          & 0 \\
CodeGen2 (original)               & 0           & 0             & 0           & 0.64\%  \\
CodeGen2.5 (original)             & 17.24\%     & 23.08\%       & 17.24\%           & 24.36\% \\
% DeepSeek-Coder-V2-lite (original) & 44.83\%     & 50\%             & 44.83\%           & -             & -           & -            \\
% DeepSeek-V2-lite (original)       & 17.24\%     & 26.28\%        & 17.24\%           & -             & -           & -            \\ 
\midrule
CodeLLama (7b)                    & 6.90\%      & 1.92\%        & 6.90\%      & 6.41\% \\
CodeT5+ (220m-bimodal)            & 3.45\%      & 4.49\%        & 3.45\%      & 4.49\% \\
CodeGen2 (1b)                     & 13.79\%     & 10.90\%             & 13.79\%          & 10.90\%     \\
CodeV (QW 7b) \cite{zhao2024codev}                   & 14.80\%*      & 4.5\%*      & 14.80\%*      & 4.5\%* \\
CodeGen2.5 (7b)                   & \textbf{24.14}\%     & \textbf{24.36}\%       & \textbf{24.14}\%           & \textbf{25\%}\\
% DeepSeek-Coder-V2-lite (16b)      & 24.14\%     & 37.04\%             & 34.48\%           & -             & -           & -            \\
% DeepSeek-V2-lite (16b)            & 13.79\%     &16.03\%             & 13.79\%           & -             & -           & -            \\ 
\bottomrule
\end{tabular}

\label{tab:exper_gen_com}
\end{table}

% \begin{table}[ht]
% \centering
% \begin{tabular}{lccc}
% \hline
% \textbf{Based LLM} & \textbf{Pass@1} & \textbf{Pass@5} & \textbf{Pass@10} \\ 
% CodeT5+ (original)      \\ 
% \hline
% LLama2 (7b)       \\
% CodeLLama (7b)      \\ 
% CodeT5+ (220m)     \\ 
% CodeT5+ (6b)       \\ 
% CodeGen2 (1b)      \\ 
% CodeGen2.5 (7b)        \\ 
% DeepSeek-Coder-V2-lite (16b)        \\ 
% DeepSeek-V2-lite (16b)       \\ 
% \hline
% \end{tabular}
% \caption{Pass@K}
% \label{tab:dataset_repo_levels}
% \end{table}

\subsection{PPA Prediction}
\label{sec:PPA}
\subsubsection{Overview}
Optimizing Power, Performance, and Area (PPA) stands out as a primary objective in the circuit design process. Estimating PPA metrics at an early stage not only boosts design efficiency but also empowers a more agile response to evolving design requirements and constraints. Unlike previous RTL datasets that solely gather Verilog files, our datasets encompass entire repositories, allowing us to obtain post-mapping netlists and logic synthesis reports by EDA tools. Consequently, we can parse the PPA metrics from the logic synthesis reports and predict them for the corresponding RTL designs at an early stage. In this subsection, we formulate the PPA prediction task within our dataset and assess the effectiveness of current learning-based prediction models~\cite{xu2022sns, sengupta2022good, fang2023masterrtl}. 

\subsubsection{Evaluation Metrics}
We regard the circuit charateristics of the post-mapping netlist produced by the Design Compiler tool with default settings as the ground truth for PPA metrics. Specifically, we define the dynamic power under random workloads as the \textbf{Power} metric, the maximum path delay as the \textbf{Delay} metric and the total cell area as the \textbf{Area} metric. 

We use Mean Absolute Error Percentage (MAPE) and Root Relative Square Error (RRSE) to evaluate the prediction accuracy of PPA.
It should be denoted that the PPA predictor trained on the datasets generated with a certain technology library cannot be generalized to another, as the PPA metrics of netlists are strongly linked to the technology library. Therefore, we choose the netlists and logic synthesis reports with a specific library, which keeps consistency with the settings in~\cite{xu2022sns, sengupta2022good, fang2023masterrtl}. We opt for skywater 130nm library in the following experiments.

\subsubsection{Analysis}
We collect 146 RTL designs for training and 10 designs for testing, with these designs containing more than 10k cells after logic synthesis and closely resembling practical designs. To explore the data scalability of PPA prediction models, we sample the training dataset size to 10\%, 50\% and 100\%. It should be denoted that we exclude the model~\cite{sengupta2022good} for the delay prediction as it cannot support this task. 
According to the prediction results shown in Table~\ref{TAB:PPA}, we conclude three observations. 
Firstly, the accuracy of predictions improves as the training data volume increases. For instance, the MAPE of area prediction is 4.3201 with 10\% data and decreases to 0.3303 with 100\% data.
Secondly, PPA predictors demonstrate weaker performance in delay prediction. For example, both models exhibit unsatisfactory performance, where~\cite{xu2022sns} shows 4.7392 MAPE and~\cite{fang2023masterrtl} has 3.4769 MAPE. This suggests that estimating timing characteristics in the early stages using path features on RTL-based graphs is still difficult, as logic synthesis tools heavily optimize logic to minimize maximum path delays.
Thirdly, in comparison to the originally reported performance on simple benchmarks, these models exhibit diminished performance on designs more than 10k cells in our dataset. 
Therefore, how to accurately predict PPA of practical designs remains an opening question, necessitating further exploration in the EDA community. 

\begin{table}[]
\caption{PPA prediction results of learning-based models (the lower, the better).}
\vspace{1em}
\resizebox{\linewidth}{!}{
\begin{tabular}{@{}c|l|ll|ll|ll@{}} \toprule
\multirow{2}{*}{Model}            & \multicolumn{1}{c|}{\multirow{2}{*}{Data}} & \multicolumn{2}{c|}{Area} & \multicolumn{2}{c|}{Power} & \multicolumn{2}{c}{Delay} \\ 
                                  & \multicolumn{1}{c|}{}                      & MAPE        & RRSE        & MAPE         & RRSE        & MAPE         & RRSE       \\ \midrule
\multirow{3}{*}{SNS\cite{xu2022sns}}              & 10\%                                       & 1.9379      & 1.4913      & 1.9045       & 1.1362      & 49.3638      & 6.6901     \\
                                  & 50\%                                       & 0.7811      & 1.9041      & 0.8425       & 1.1759      & 47.7314      & 2.8816     \\
                                  & 100\%                                      & 0.6564      & 1.7197      & 0.7549       & 1.1740      & 4.7392       & \textbf{2.3647}     \\ \midrule
\multirow{3}{*}{\cite{sengupta2022good}} & 10\%                                       & 1.9066      & 1.0802      & 10.2783      & 14.7510     & -            & -          \\
                                  & 50\%                                       & 0.7954      & 0.6133      & 0.7478       & \textbf{0.3539}      & -            & -          \\
                                  & 100\%                                      & 0.5996      & \textbf{0.5074}      & 0.7343       & 0.6558      & -            & -          \\ \midrule
\multirow{3}{*}{MasterRTL\cite{fang2023masterrtl}}        & 10\%                                       & 1.3405      & 1.1197      & 2.3253       & 29.7767     & 80.7779      & 8.5936     \\
                                  & 50\%                                       & 0.5640      & 0.9982      & 1.7237       & 29.8749     & 24.7716      & 6.8725     \\
                                  & 100\%                                      & \textbf{0.3303}      & 0.7605      & \textbf{0.6501}       & 2.2541      & \textbf{3.4769}       & 4.2863    \\ \bottomrule
\end{tabular}}

\label{TAB:PPA}
\end{table}

\section{Conclusion}\label{Sec:Conclusion}

In this paper, we introduce \textbf{DeepCircuitX}, a multimodal and comprehensive repository-level dataset designed to advance RTL code understanding, generation, and PPA analysis in hardware design automation. By offering a holistic resource that spans repository, file, module, and block-level RTL code, DeepCircuitX enables large language models to better tackle the complexities of hardware design. The integration of Chain of Thought annotations further enhances the dataset's value by providing detailed insights into functionality and structure, thereby improving model training and performance across a variety of RTL tasks.
Our experiments demonstrate that models trained on DeepCircuitX significantly outperform existing methods in tasks like code understanding, generation, and completion, as well as in RTL-to-PPA prediction. The inclusion of synthesized netlists and PPA metrics opens up new avenues for early-stage design exploration, offering a practical tool for both researchers and practitioners in EDA. As a result, DeepCircuitX establishes new benchmarks for RTL tasks, setting a foundation for future innovations in hardware design automation and demonstrating the potential of LLMs to transform this critical domain.

\balance

% \section*{Acknowledgments}

% \newpage
\bibliographystyle{IEEEtran}
\bibliography{reference}

\end{document}